\def\year{2020}\relax
\documentclass[letterpaper]{article} % DO NOT CHANGE THIS
\usepackage{aaai20}  % DO NOT CHANGE THIS
\usepackage{times}  % DO NOT CHANGE THIS
\usepackage{helvet} % DO NOT CHANGE THIS
\usepackage{courier}  % DO NOT CHANGE THIS
\usepackage[hyphens]{url}  % DO NOT CHANGE THIS
\usepackage{graphicx} % DO NOT CHANGE THIS
\usepackage{graphicx}  % DO NOT CHANGE THIS
\usepackage{amsmath}
\usepackage{amssymb}
\usepackage{booktabs}
\newcommand{\beginsupplement}{%
    \setcounter{table}{0}
    \renewcommand{\thetable}{S\arabic{table}}%
    \setcounter{figure}{0}
    \renewcommand{\thefigure}{S\arabic{figure}}%
    \setcounter{section}{0}
    \renewcommand{\thesection}{S}%
}

\title{Detecting Human-Object Interactions via Functional Generalization}
\author{Ankan Bansal, Sai Saketh Rambhatla, Abhinav Shrivastava, Rama Chellappa \\ 
        University of Maryland, College Park\\
    \{ankan,rssaketh,abhinav,rama\}@umiacs.umd.edu}

\begin{document}

\maketitle

%%%%%%%%%%%%%%%%%%%%%%%%%%%%%%%%%%%%%%%% ABSTRACT%%%%%%%%%%%%%%%%%%%%%%%%%%%%%%%%%%%%%%%%%%%%%%
\begin{abstract}
    We present an approach for detecting human-object interactions (HOIs) in images, based on the
    idea that humans interact with functionally similar objects in a similar manner. The proposed
    model is simple and efficiently uses the data, visual features of the human, relative
    spatial orientation of the
    human and the object, and the knowledge that functionally similar objects take part in similar
    interactions with humans. We provide extensive experimental validation for our approach and
    demonstrate state-of-the-art results for HOI detection. On the HICO-Det dataset our method
    achieves a gain of over $2.5\%$ absolute points in mean average precision (mAP) over
    state-of-the-art. We also show
    that our approach leads to significant performance gains for zero-shot HOI detection in the seen
    object setting. We further demonstrate that using a generic object detector, our model can
    generalize to interactions involving previously unseen objects. 
\end{abstract}

%%%%%%%%%%%%%%%%%%%%%%%%%%%%%%%%%%%%INTRODUCTION%%%%%%%%%%%%%%%%%%%%%%%%%%%%%%%%%%%%%%%%%%%%%%%
\section{Introduction}
Human-object interaction (HOI) detection is the task of localizing and inferring relationships
between a human and an object, e.g., ``eating an apple'' or ``riding a bike.'' Given an input image, the
standard representation for HOIs \cite{sadeghi2011recognition,gupta2015visual} is a triplet
$\langle$\texttt{human}, \texttt{predicate}, \texttt{object}$\rangle$, where \texttt{human} and
\texttt{object} are represented by bounding boxes, and \texttt{predicate} is the interaction between
this $($\texttt{human}, \texttt{object}$)$ pair. At first glance, it seems that this problem is a
composition of the atomic problems of human and object detection 
and HOI classification \cite{shen2018scaling,gkioxari2017detecting}. These atomic recognition tasks
are certainly the building blocks of a variety of
approaches for HOI understanding \cite{shen2018scaling,delaitre2011learning}; and the progress in
these atomic tasks directly translates to improvements in HOI understanding. However, the task of
HOI understanding comes with its own unique set of challenges
\cite{lu2016visual,chao2017learning}.

\begin{figure}[t]
   \centering
   \includegraphics[scale=0.36]{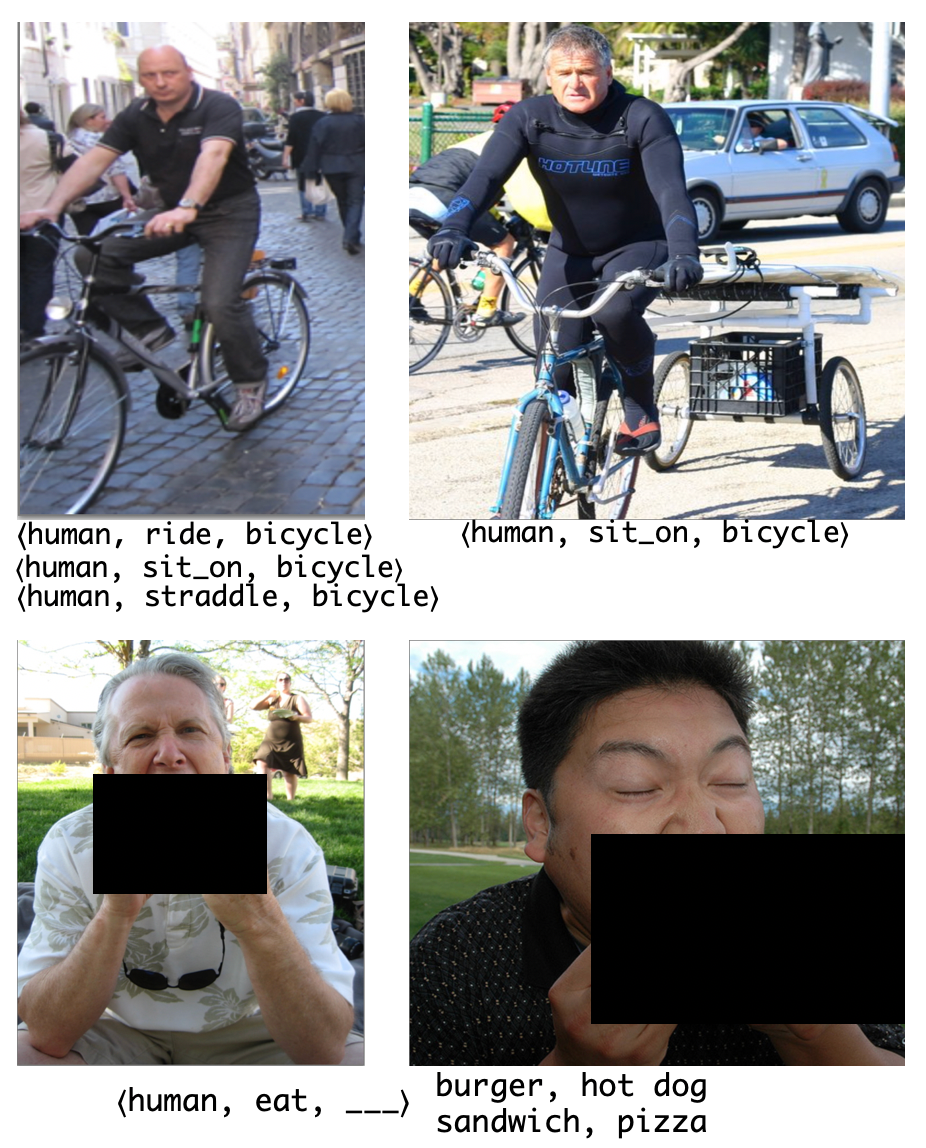}
   \caption{Common properties of HOI Detection. Top - Datasets are not
       exhaustively labeled. Bottom - Humans interact similarly with functionally
   similar objects - both persons could be eating either a burger, a hot dog, or a
   pizza.}
\label{fig:intro}
\end{figure}

These challenges are due to the combinatorial explosion of the possible interactions with increasing
number of objects and predicates. For example, in the commonly used HICO-Det dataset
\cite{chao2017learning} with 80 unique object classes and 117 predicates, there are 9,360 possible
relationships. This number increases to more than $10^6$ for larger datasets like Visual Genome
\cite{krishna2017visual} and HCVRD \cite{zhuang2017care}, which have hundreds of object categories
and thousands of predicates. This, combined with the long-tail distribution of HOI categories, makes
it difficult to collect labeled training data for all HOI triplets. A common solution to this
problem is to arbitrarily limit the set of HOI relationships and only collect labeled images for
this limited subset. For example, the HICO-Det benchmark has only 600 unique relationships.

Though these datasets can be used for training models for recognizing a
limited set of HOI triplets, they do not address the problem completely. For example, consider the
images shown in Figure \ref{fig:intro} (top row) from the challenging HICO-Det dataset. The three
pseudo-synonymous relationships: $\langle$\texttt{human}, \texttt{hold}, \texttt{bicycle}$\rangle$,
$\langle$\texttt{human}, \texttt{sit$\_$on}, \texttt{bicycle}$\rangle$, and $\langle$\texttt{human},
\texttt{straddle}, \texttt{bicycle}$\rangle$ are all possible for both these images; but only a
subset is labeled for each. We argue that this is not a quality control issue while collecting a
dataset, but a problem associated with the huge space of possible HOI relationships. It is
enormously challenging to exhaustively label even the 600 unique HOIs, let alone all possible
interactions between humans and objects. An HOI detection model that relies entirely on labeled data
will be unable to recognize the relationship triplets that are not present in the dataset, but are
common in the real-world. For example, a na\"{\i}ve model trained on HICO-Det cannot recognize the 
$\langle$\texttt{human}, \texttt{push}, \texttt{car}$\rangle$ triplet because this triplet does not
exist in the training set. The ability to recognize previously unseen relationships (zero-shot
recognition) is a highly desirable capability for HOI detection.

In this work, we address the challenges discussed above using a model that leverages the
common-sense knowledge that humans have similar interactions with objects that are functionally
similar. The proposed model can inherently do zero-shot detection. Consider the images
in Figure \ref{fig:intro} (second row) with $\langle$\texttt{human}, \texttt{eat}, \texttt{?}$\rangle$
triplet. The person in either image could be eating a burger, a sandwich, a hot dog, or a
pizza. Inspired by this, our key contribution is incorporating this common-sense knowledge in a model for
generalizing HOI detection to functionally similar objects. This model utilizes visual appearance of
a human, their relative geometry with the object, and language
priors~\cite{Mikolov2013DistributedRO} to capture which objects afford similar predicates
\cite{gibson1979theory}. Such a model is able to exploit the large amount of contextual information
present in the language priors to generalize HOIs across functionally similar objects.

In order to train this module, we need a list of functionally similar
objects and labeled examples for the relevant HOI triplets, neither of which are readily available.
To overcome this, we propose a way to train this model by: 1) using a large vocabulary of objects,
2) discovering functionally similar objects automatically, and 3) proposing data-augmentation,
emulating the examples shown in Figure \ref{fig:intro} (second row). 
To discover functionally similar objects in an unsupervised way, we use a combination of visual
appearance features and semantic word embeddings~\cite{Mikolov2013DistributedRO}
to represent the objects in a ``world set" (Open Images Dataset (OID)~\cite{OpenImages}). Note that
the proposed method is not contingent on the world set. Any large dataset, like
ImageNet, could replace OID. Finally, to emulate the
examples shown in Figure \ref{fig:intro} (second row), we use the human and object bounding boxes from
a labeled interaction, the visual features from the human bounding box, and semantic word embeddings
of all functionally similar objects. Notice that this step does not utilize the visual features for
objects, just their relative locations with respect to a human, enabling us to perform this
data-augmentation. Further, to efficiently use the training data, we fine-tune the object detector
on the HICO-Det dataset unlike prior approaches. 

The proposed approach achieves over $2.5\%$ absolute improvement in mAP over the best published method for
HICO-Det. Further, using a generic object detector, and the proposed functional generalization model lends itself
directly to the zero-shot HOI triplet detection problem. We clarify that zero-shot detection is the
problem of detecting HOI triplets for which the model has never seen any images. Knowledge about
functionally similar objects enables our system to detect interactions involving objects not
contained in the original training set. Using just this generic object detector, our model achieves
state-of-the-art performance for HOI detection on the popular HICO-Det dataset in the zero-shot
setting, improving over existing methods by several percentage points. Additionally, we show that
the proposed approach can be used as a way to deal with social/systematic biases present in
vision$+$language datasets \cite{zhao2017men,anne2018women}.

In summary, the contributions of this paper are: (1) a functional generalization model for capturing
functional similarities between objects; (2) a method for training the proposed model; and (3)
state-of-the-art results on HICO-Det in both fully-supervised and zero-shot settings.%; and (4) a
%filtered version of HCVRD and a strong zero-shot HOI detection baseline.

%%%%%%%%%%%%%%%%%%%%%%%%%%%%%%%%%%%%%%%%RELATED%%%%%%%%%%%%%%%%%%%%%%%%%%%%%%%%%%%%%%%%%%%%%%%%%%
\section{Related Work}
\label{sec:related}

\textbf{Human-Object Interaction.}
Early methods \cite{yao2010grouplet,yao2011human} relied on structured visual
features which capture contextual relationships between humans and objects. Similarly,
\cite{delaitre2011learning} used structured representations and spatial co-occurrences of body
parts and objects to train models for HOI recognition. Gupta \emph{et al.}
\cite{gupta2007objects,gupta2009observing} adopted a Bayesian approach that integrated object
classification and localization, action understanding, and perception of object reaction.
\cite{desai2012detecting} constructed a compositional model which combined skeleton models,
    poselets, and visual phrases.

More recently, with the release of large datasets like HICO \cite{chao2015hico}, Visual Genome
\cite{krishna2017visual}, HCVRD \cite{zhuang2017care}, V-COCO \cite{gupta2015visual}, and HICO-Det
\cite{chao2017learning}, the problem of detecting and recognizing HOIs has attracted significant
attention. This has been driven by HICO which is a benchmark dataset for recognizing human-object
interactions. The HICO-Det dataset extended HICO by adding bounding box annotations. V-COCO is a
much smaller dataset containing 26 classes and about 10,000 images. On the other hand, HCVRD and
Visual Genome provide annotations for thousands of relationship categories and hundreds of objects.
However, they suffer from noisy labels. We primarily use the HICO-Det dataset to evaluate our
approach in this paper.

\cite{gkioxari2017detecting} designed a system which trains object and relationship detectors
simultaneously on the same dataset and classifies a human-object pair into a fixed set of
pre-defined relationship classes. This precludes the method from being useful for detecting novel
relationships. \cite{xu2018interact} used pose and gaze information for HOI detection.
\cite{kolesnikov2018detecting}
introduced the Box Attention module to a standard R-CNN and trained simultaneously for object
detection and relationship triplet prediction. Graph Parsing Neural Networks \cite{qi2018learning}
incorporated structural knowledge and inferred a parse graph in a message passing inference
framework. In contrast, our method does not need iterative processing and requires only a single
pass through a neural network.

Unlike most prior work, we do not directly classify into a fixed set of
relationship triplets but into predicates. This helps us detect previously unseen interactions. The
method closest in spirit to our approach is \cite{shen2018scaling} which uses a
two branch structure with the first branch responsible for detecting humans and 
predicates, and the second for detecting objects. Unlike our proposed approach, their method
 solely depends on the appearance of the
human. Also, they do not use any prior information from language. Our model utilizes implicit human
appearance, the object label, human-object geometric relationship, and knowledge about
similarities between objects. Hence, our model achieves much better performance than \cite{shen2018scaling}.

We also distinguish our work from prior work \cite{kato2018compositional,fang2018pairwise} on HOI
recognition. 
We tackle the more difficult problem of detecting HOIs here. 

\textbf{Zero-shot Learning.} Our work also ties well with zero-shot classification
\cite{xian2017zero,kodirov2017semantic} and zero-shot object detection (ZSD)
    \cite{bansal2018zero}. \cite{bansal2018zero}
    proposed projecting images into the word-vector space to exploit the semantic properties of such
    spaces. They also discussed challenges associated with training and evaluating
    ZSD. A similar idea was used in \cite{kodirov2017semantic} for zero-shot classification.
    \cite{rahman2018zero}, on the other hand, used meta-classes to cluster semantically
    similar classes. In this work, we also use word-vectors
    as semantic information for our generalization module. This, along with our approach for
    generalization during training, helps zero-shot HOI detection.

%%%%%%%%%%%%%%%%%%%%%%%%%%%%%%%%%%%%%%%%%%%%%APPROACH%%%%%%%%%%%%%%%%%%%%%%%%%%%%%%%%%%%%%%%%%%%%%
\section{Approach}
\label{sec:approach}

\begin{figure*}[ht]
    \centering
        \includegraphics[scale=0.52]{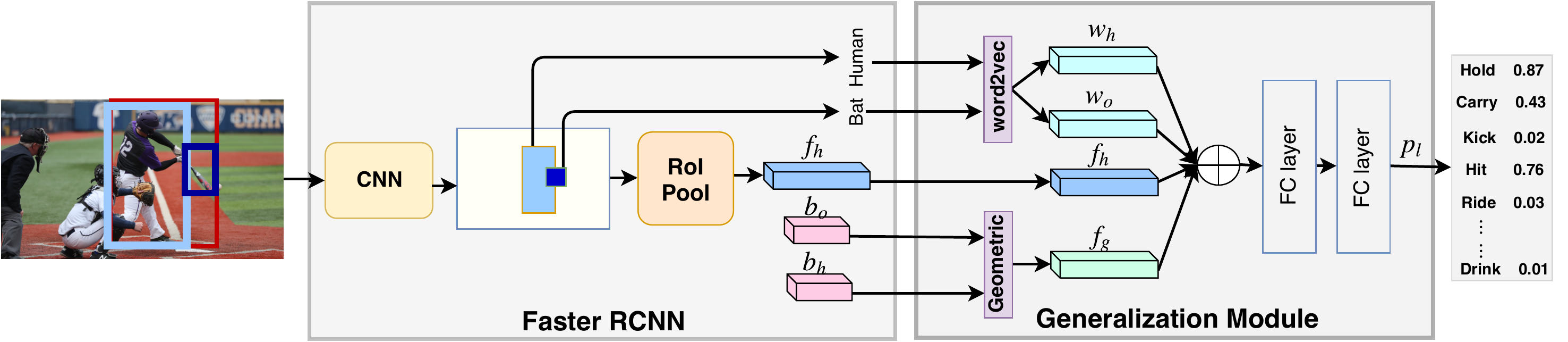}
    \caption{We detect all objects and humans in an image. This detector gives human features $f_h$, and
    the corresponding labels. We consider all pairs of human-object and create union boxes. Our
    functional generalization module uses the word vectors for the human $w_h$, the object class
    $w_o$, geometric features $f_g$, and $f_h$ to produce the probability estimate over the
predicates.}
    \label{fig:approach}
\end{figure*}

Figure \ref{fig:approach} represents our approach. The main novelty of our proposed approach lies 
in incorporating generalization through a language component. This is done by using functional
similarities of objects during training. For inference, we first detect humans and objects in the
image using our object detectors, which also give the corresponding
(RoI-pooled~\cite{ren2015faster}) feature representations. Each human-object pair is used
to extract visual and language features which are used to predict the predicate associated with the
interaction. We describe each component of the model and the training procedure in the
following sections.

\subsection{Object Detection}
In the fully-supervised setting, we use an object detector fine-tuned on the
HICO-Det dataset. For zero-shot detection and further experiments, we use a Faster-RCNN
\cite{huang2017speed} based detector trained on the Open Images dataset (OID) \cite{OpenImages}.
This network can detect 545 object categories and we use it to obtain proposals for humans and
objects in an image. The object detectors also output the ROI-pooled features
corresponding to these detections. All human-object pairs thus obtained are passed to our model
which outputs probabilities for each predicate.

\subsection{Functional Generalization Module}
Humans look similar when they interact with functionally similar objects. Leveraging this fact, the
functional generalization module exploits object similarities, the relative spatial location of
human and object boxes, and the implicit human appearance to estimate the predicate. At its core, it
comprises a Multi Layer Perceptron (MLP), which takes as input the human and object word embeddings,
$w_h$ and $w_o$, the geometric relationship between the human and object boxes $f_g$, and the human
visual feature $f_h$. The geometric feature is useful as the relative positions of a human and an
object can help eliminate certain predicates. The human feature $f_h$ is used as a representation
for the appearance of the human. This appearance representation is added because the aim is to
incorporate the idea that humans look similar while interacting with similar objects. For example, a
person drinking from a cup looks similar while drinking from a glass or a bottle. The four features
$w_h$, $w_o$, $f_g$, and $f_h$ are concatenated and passed through a 2-layer MLP which predicts
the probabilities for each predicate. All the predicates are considered independent. We now give
details of different components in this model.

\subsubsection{Word embeddings.} 
We use 300-D vectors from word2vec \cite{Mikolov2013DistributedRO} to get the human and object
embeddings $w_h$ and $w_o$. Object embeddings allow discovery of previously unseen interactions by
exploiting semantic similarities between objects. The human embedding, $w_h$, helps in
distinguishing between different words for humans (man/woman/boy/girl/person), if required.

\subsubsection{Geometric features.} 
Following prior work on visual relationship detection \cite{zhuang2017towards}, we define the
geometric relationship feature as:
\begin{multline}
f_g = \left[\frac{x_1^h}{W}, \frac{y_1^h}{H}, \frac{x_2^h}{W}, \frac{y_2^h}{H}, \frac{A^h}{A^I}, \frac{x_1^o}{W}, \frac{y_1^o}{H}, \frac{x_2^o}{W}, \frac{y_2^o}{H}, \frac{A^o}{A^I}, \right. \\
\left. \left(\frac{x_1^h - x_1^o}{x_2^o - x_1^o}\right), \left(\frac{y_1^h - y_1^o}{y_2^o - y_1^o}\right), \right. \\ 
\left. \log\left(\frac{x_2^h - x_1^h}{x_2^o - x_1^o}\right), \log\left(\frac{y_2^h - y_1^h}{y_2^o - y_1^o}\right) \right]
\end{multline}
where, $W, H$ are the image width and height, $(x_i^h, y_i^h)$, and $(x_i^o, y_i^o)$ are the human
and object bounding box coordinates respectively, $A^h$ is the area of the human box, $A^o$ is the
area of the object box, and $A^I$ is the area of the image. The geometric feature $f_g$ uses spatial
features for both entities (human and object) and also spatial features from their relationship. It
encodes the relative positions of the two entities.

\subsubsection{Generalizing to new HOIs.} 
\begin{figure}
    \centering
    \includegraphics[scale=0.5]{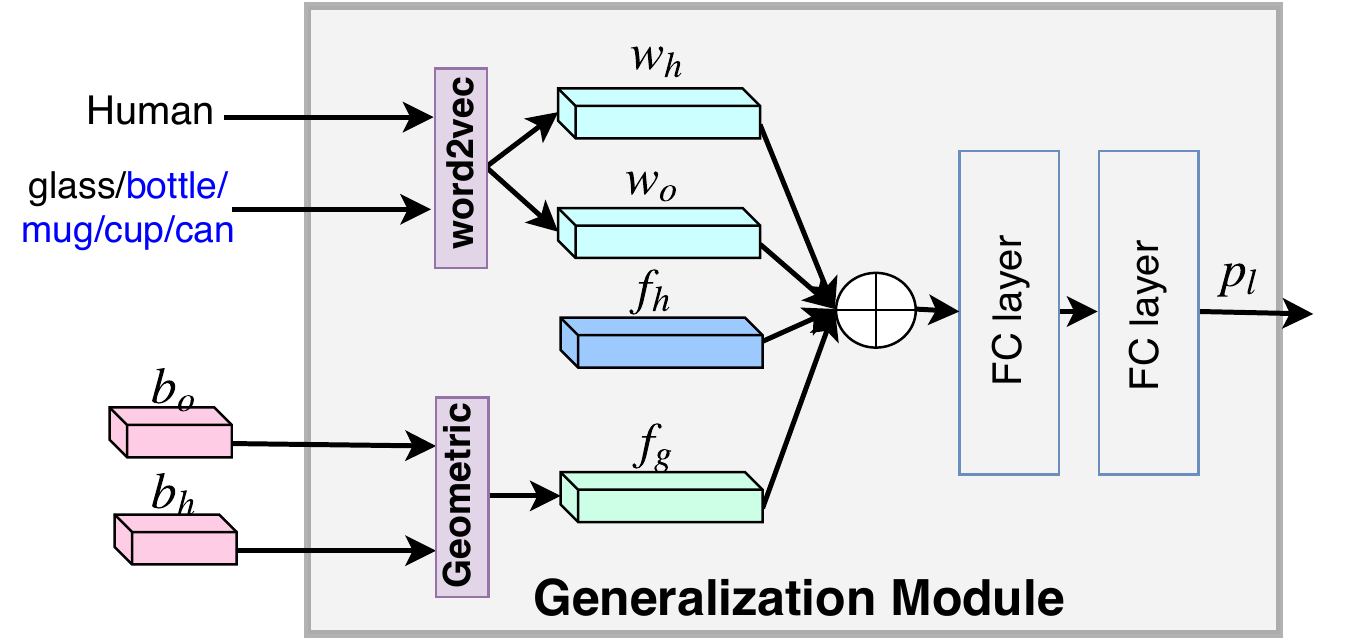}
    \caption{Generalization module. During training, we can replace ``glass" by ``bottle", ``mug", ``cup", or ``can".}
    \label{fig:c_sense}
\end{figure}

We incorporate the idea that humans interacting with similar objects look similar via the
functional generalization module. As shown in figure \ref{fig:c_sense}, this idea can be added by
changing the object name while keeping the human word vector $w_h$, the human visual feature $f_h$,
and the geometric feature $f_g$ fixed. Each object has a different word-vector and the model learns
to recognize the same predicate for different human-object pairs. Note that this does not need
visual examples for all human-object pairs.

\noindent
\textbf{Finding similar objects.} A na\"{i}ve choice for defining similarity between objects
would be through the WordNet hierarchy \cite{miller1995wordnet}. However, several issues make using
WordNet impractical. The first is defining
distance between the nodes in the tree. The height of a node cannot be used as a metric because
different things have different levels of categorization in the tree. Similarly, defining sibling
relationships which adhere to functional intuitions is challenging. Another issue
is the lack of correspondence between closeness in the tree and semantic
similarities between objects.

To overcome these problems, we consider similarity in both the visual and semantic representations
of objects. We start by defining a vocabulary of objects $\mathcal{V} = \{o_1, \dots, o_n\}$ which
includes all the objects that can be detected by our object detector. For each object $o_i \in
\mathcal{V}$, we obtain a visual feature $f_{o_i}\in \mathbb{R}^p$ from images in OID, and a word
vector $w_{o_i} \in \mathbb{R}^q$. We concatenate these two to obtain the mixed representation
$u_{o_i}$ for object $o_i$. We then cluster $u_i$'s into $K$ clusters using Euclidean distance.
Objects in the same cluster are considered functionally similar. This clustering has to be done only
once. We use these clusters to find all objects similar to an object in the target dataset. Note
that there might not be any visual examples for many of the objects obtained using this method. 
This is why we do not use the RoI-pooled visual features from the object. 

Using either just the word2vec representations or just the visual
representations for clustering gave several inconsistent clusters. Therefore, we use the
concatenated features $u_{o_i}$. We observed that clusters created using these features better
correspond to functional similarities between objects.

\noindent
\textbf{Generating training data}. For each relationship triplet \emph{$<$h,p,o$>$} in the original
dataset, we add \emph{r} triplets \emph{$<$h,p,o$_{1}$$>$}, \emph{$<$h,p,o$_{2}$$>$}, ...,
\emph{$<$h,p,o$_r$$>$} to the dataset keeping the human, and object boxes fixed, and only changing
the object name. This means that, for all these $f_g$ and $f_h$ are the same as for the original
sample. The \emph{r} different objects, \emph{o$_1$,..., o$_r$} belong to the same cluster as object
\emph{o}. For example, in figure \ref{fig:c_sense}, the ground truth category ``glass" can be
replaced by ``bottle", ``mug", ``cup", or ``can" while keeping $w_h$, $f_h$, and $f_g$ fixed.

\subsection{Training}
A training batch consists of $T$ interaction triplets. The model produces probabilities for each
predicate independently. We use a weighted class-wise BCE loss for training the
model.

\noindent\textbf{Noisy labeling}. Missing and incorrect labels are a common issue in HOI datasets.
Also, a human-object pair can have different types of interactions at the same time. For example, a
person can be sitting on a bicycle, riding a bicycle, and straddling a bicycle. These interactions
are usually labeled with slightly different bounding boxes. To overcome these issues, we use a
per-triplet loss weighing strategy. A training triplet in our dataset has a single label, e.g.
\texttt{$<$human-ride-bicycle$>$}. A triplet with slightly shifted bounding boxes might have another
label, like \texttt{$<$human-sit\_on-bicycle$>$}. The idea is that the models should be penalized more
if they fail to predict the correct class for a triplet. Given the training sample
\texttt{$<$human-ride-bicycle$>$}, we want the model to definitely predict ``\texttt{ride}", but we should not
penalize it for predicting ``\texttt{sit\_on}" as well. Therefore, while training the model,
we use the following weighing scheme for classes. Suppose that a training triplet is labeled
\texttt{$<$human-ride-bicycle$>$} and there are some other triplets in the image. For this training
triplet, we assign a high weight (10.0 here) to the loss for the correct class (\texttt{ride}), and a zero
weight to all other predicates in the image. We also scale down the weight (1.0 here) to the loss
for all other classes to ensure that the model is not penalized too much for predicting a missing
but correct label.

\subsection{Inference}
The inference step is simply a forward pass through the network (figure \ref{fig:approach}).
The final step of inference is class-wise non-maximal suppression (NMS) over the union of human and
object boxes. This helps in removing multiple detections for the same interaction and leads to
higher precisions.

%%%%%%%%%%%%%%%%%%%%%%%%%%%%%%%%%%%%%%%%%%EXPERIMENTS%%%%%%%%%%%%%%%%%%%%%%%%%%%%%%%%%%%%%%%%%%%%%%%
\section{Experiments}
\label{sec:experiments}

We evaluate our approach on the HICO-Det dataset \cite{chao2015hico}. As mentioned
before, V-COCO \cite{gupta2015visual} is a small dataset and does not provide any insights into the
proposed method. In line with recent work \cite{gupta2018no}, we avoid
using it.% to leave space for analysis of the proposed method.

\subsection{Dataset and Evaluation Metrics}
\noindent
HICO-Det extends the HICO dataset \cite{chao2015hico} which contains 600 HOI categories for 80
objects. HICO-Det adds bounding box
annotations for humans, and objects for each HOI category. The training set contains
over 38,000 images and about 120,000 HOI annotations for the 600 HOI classes. The test set has 33,400 HOI instances.

We use mean average precision (mAP) commonly used in object detection. An HOI detection is
considered a true positive if the minimum of human overlap IOU$_h$ and object overlap IOU$_o$ with
the ground truths is greater than 0.5. Performance is usually reported for three different HOI category
sets: (a) all 600 classes (Full), (b) 138 classes with less than 10 training samples
(Rare), and (c) the remaining 462 classes with more than 10 training samples (Non-Rare).

\subsection{Implementation Details}
We start with a ResNet-101 backbone Faster-RCNN which is fine-tuned for the HICO-Det dataset. This
detector was originally trained on COCO \cite{lin2014microsoft} which has the same $80$ object
categories as HICO-Det. We consider all detections for which the detection confidence is greater
than $0.9$ and create human-object pairs for each image. Each detection has an associated feature
vector. These pairs are then passed through our model. The human feature $f_h$ is $2048$
dimensional. The two hidden layers in the model are of dimensions $1024$ and $512$. The model
outputs probability estimates for each predicate independently and the final output prediction is
all predicates with probability $\geq 0.5$. We report performance with the COCO
detector in supplementary. 

For all the experiments, we train the model for 25 epochs with 0.1 initial learning rate which is
dropped by a tenth every 10 epochs. We re-iterate that the object detector and the
word2vec vectors are frozen while training this model. For all experiments we use up to five ($r$)
additional objects for augmentation, i.e., for each human-object pair in the training set,
we add up to five objects from the same cluster while leaving the bounding boxes and human features
unchanged.

\subsection{Results}
With no functional generalization, our baseline model achieves an mAP of $12.17\%$ for Rare classes
which is already higher than all but the most recent methods. This is because of a more efficient
use of the training data by using a fine-tuned object detector. The last row in table
\ref{tab:hico_results} shows the results attained by our complete model (with functional
generalization). For the Full set, it achieves over $2.5\%$ absolute improvement over the best
published work \cite{peyre2018detecting}. Our model also gives an mAP of $16.43\%$ for Rare classes
compared to the existing best of $15.65\%$ \cite{Wan_2019_ICCV}. The performance, along with
the simplicity, of our model is a remarkable strength and reveals that existing methods may be
over-engineered.

\begin{table}
\caption{mAPs (\%) in the default setting for the HICO-Det dataset. Our model was trained with
up to five neighbors. The last column is the total number of parameters in the
proposed classification models.
}
    \begin{center}
        \resizebox{\linewidth}{!}
        {
        \begin{tabular}{@{}lccc|c@{}}
        \toprule
        & \textbf{Full} & \textbf{Rare} & \textbf{Non-Rare} & \textbf{Params.}\\
        \textbf{Method} & (600) & (138) & (462) & (millions)\\
        \midrule
        \cite{shen2018scaling} & 6.46 & 4.24 & 7.12 & -\\
        %\cite{chao2017learning} & 7.30 & 4.68 & 8.08 & -\\
        \cite{chao2017learning} & 7.81 & 5.37 & 8.54 & -\\
        \cite{gkioxari2017detecting} & 9.94 & 7.16 & 10.77 & -\\
        \cite{xu2018interact} & 9.97 & 7.11 & 10.83 & -\\
        \cite{qi2018learning} & 13.11 & 9.34 & 14.23 & -\\
        \cite{Xu_2019_CVPR} & 14.70 & 13.26 & 15.13 & - \\
        \cite{gao2018ican} & 14.84 & 10.45 & 16.15 & 48.1 \\
        \cite{Wang_2019_ICCV} & 16.24 & 11.16 & 17.75 & - \\
        \cite{gupta2018no} & 17.18 & 12.17 & 18.68 & 9.2\\
        \cite{li2018transferable} & 17.22 & 13.51 & 18.32 & 35.0\\
        \cite{Zhou_2019_ICCV} & 17.35 & 12.78 & 18.71 & - \\
        \cite{Wan_2019_ICCV} & 17.46 & 15.65 & 18.00 & - \\
        \cite{peyre2018detecting} & 19.40 & 15.40 & 20.75 & 21.8\\ %New
        \midrule
        Ours & \textbf{21.96} & \textbf{16.43} & \textbf{23.62} & \textbf{3.1} \\
        \bottomrule
        \end{tabular}
        }
    \end{center}
        \label{tab:hico_results}
\end{table}

\subsubsection{Comparison of number of parameters.}
In table \ref{tab:hico_results}, we also compare the number of parameters in four recent
models against our model. With far fewer parameters, our model achieves better performance.
For example, compared to the current state-of-the-art model which contains $62.7$ million parameters
and achieves only $19.40\%$ mAP, our model contains just $51.1$ million parameters and reaches an
mAP of $21.96\%$. Ignoring the object detectors, our model introduces just $3.1$ million new
parameters.  (Due to lack of specific details in previous papers, we have made some conservative
assumptions which we list in the supplementary material.)  In addition, the approaches in
\cite{gupta2018no} and \cite{li2018transferable} require pose estimation models too. The numbers
listed in table \ref{tab:hico_results} do not count these parameters. The strength of our method is the
simple and intuitive way of thinking about the problem.

Next, we show how a generic object detector can be used to detect novel interactions, even those
involving objects not present in the training set. We will use an off-the-shelf Faster
RCNN which is trained on OpenImages and is capable of detecting
545 object categories.  This detector uses an Inception ResNet-v2 with atrous convolutions as its
base network.

\subsection{Zero-shot HOI Detection}
\cite{shen2018scaling} take the idea of zero-shot object detection further and try
to detect previously unseen human-object relationships in images. The aim is to
detect interactions for which no images are available during training. In this section, we show that
our method offers significant improvements over \cite{shen2018scaling} for zero-shot HOI detection.

\subsubsection{Seen object scenario.}
We first consider the same setting as \cite{shen2018scaling}. We select 120 relationship triplets
ensuring that every object involved in these 120 relationships occurs in at least one of the
remaining 480 triplets. We call this the ``seen object" setting, i.e., the model sees all
the objects involved but not all relationships. Later, we will introduce the ``unseen object" where
no relationships involving a set of objects will be observed during training.

Table \ref{tab:zero_shot_seen_object} shows the performance of our approach in the ``seen object"
setting for 120 unseen triplets during training. Note that, since \cite{shen2018scaling} have not
release the list of classes publicly, we report the mean 
over 5 random sets of 120 unseen classes in table \ref{tab:zero_shot_seen_object}. We achieve
significant improvement over the prior method.

\begin{table}
\caption{mAPs (\%) in the default setting for ZSD. This is the seen object setting, i.e., all the
objects have been seen.}
    \begin{center}
        \resizebox{0.42\textwidth}{!}
        {
        \begin{tabular}{@{}lccc@{}}
        \toprule
        & \textbf{Unseen} & \textbf{Seen} & \textbf{All}\\
        \textbf{Method}                 &   (120 classes)   &    (480)  &  (600) \\
        \midrule
        \cite{shen2018scaling} & 5.62 & - & 6.26 \\
        \midrule
        Ours & \textbf{11.31}$\pm$1.03 & \textbf{12.74}$\pm$0.34 & \textbf{12.45}$\pm$0.16 \\
        \bottomrule
        \end{tabular}
    }
    \end{center}
    \label{tab:zero_shot_seen_object}
\end{table}

\subsubsection{Unseen object scenario.}
We start by randomly selecting 12 objects from the 80 objects in HICO. We pick all relationships
containing these objects. This gives us 100 relationship triplets which constitute the test (unseen)
set. We train models using visual examples from only the remaining 500
categories. Table \ref{tab:zero_shot_unseen_object} gives results for our methods in this setting.
We cannot compare with existing methods because none of them have the ability to detect HOIs in the
unseen object scenario. We hope that our method will serve as a baseline for future research on this
important problem.

\begin{table}
\caption{mAPs (\%) in the unseen object setting for ZSD. This is the unseen object setting where
        the trained model for interaction recognition has not seen any examples of some object
    classes.}
    \begin{center}
        \resizebox{0.68\linewidth}{!}
        { 
        \begin{tabular}{@{}lccc@{}}
        \toprule
        & \textbf{Unseen} & \textbf{Seen} & \textbf{All}\\
        \textbf{Method} &   (100 classes)   &   (500)  &    (600) \\
        \midrule
        Ours & 11.22 & 14.36 & 13.84 \\
        \bottomrule
        \end{tabular}
        }
    \end{center}
    \label{tab:zero_shot_unseen_object}
\end{table}

\begin{figure*}[t!]
    \centering
        \includegraphics[width=0.9\linewidth]{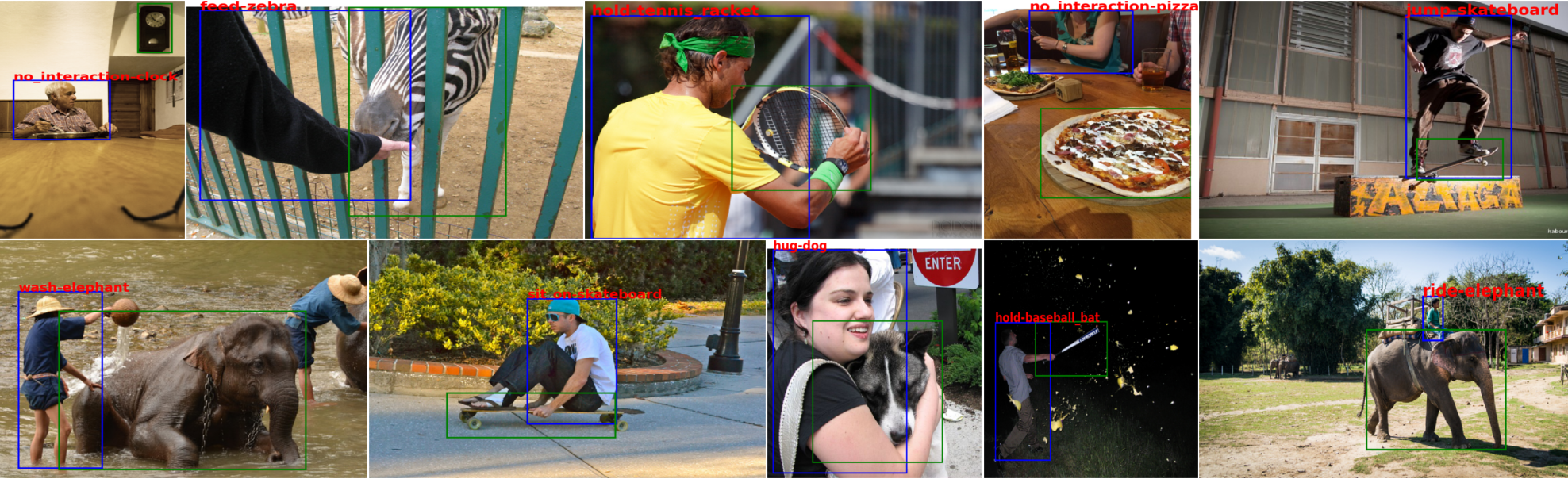}
        \caption{Some HOI detections in the unseen object ZSD setting. Our model has not seen any
        image with the objects shown above during training. (We show some mistakes made by the model
    in the supplementary material.)}
    \label{fig:unseen_examples}
\end{figure*}

In figure \ref{fig:unseen_examples}, we show that our model can detect interaction triplets with
unseen objects. This is because we use a generic detector which can detect
many more objects. We note, here, that there are some classes among the 80 COCO classes which do not
occur in OI. We willingly take the penalty for missing interactions with these objects in order
to present a more robust system which not only works for the dataset of interest but is able to
generalize to completely unseen interaction classes. We reiterate that none of the previous methods
has the ability to detect HOIs in this scenario.

\subsection{Ablation Analysis}
The generic object detector used for zero-shot HOI detection can also be used in the supervised
setting. For example, using this detector, we obtain an mAP of $14.35\%$ on the Full set of
HICO-Det. This is a competitive performance and is worse (table
\ref{tab:hico_results}) than only the most recent works. This shows the strength of generalization.  In
this section, we provide further analysis of our model with the generic detector.

\subsubsection{Number of neighbors.} Table \ref{tab:variation_with_k} shows the effect of varying
the number of neighboring objects which are added to the dataset for each training instance. The
baseline (first row) is when no additional objects are added. This is when we rely only on the
interactions present in the original dataset.  We successively add interactions with neighboring
objects to the training data and observe that the performance improves significantly. However, since
the clusters are not perfect, adding more neighbors can start becoming harmful. Also, the training
times increase rapidly. Therefore, we add five neighbors for each HOI instance in all our
experiments.

\begin{table}
\caption{HICO-Det performance (mAP \%) of the model with different number of neighbors
considered for generalization.}
    \begin{center}
        \resizebox{0.4\textwidth}{!}
        {
        \begin{tabular}{@{}cccc@{}}
        \toprule
        \textbf{r} & \textbf{Full} & \textbf{Rare} & \textbf{Non-Rare}\\
        (Number of objects) & (600 classes) & (138) & (462) \\
        \midrule
        0 & 12.72 & 7.57 & 14.26 \\
        3 & 13.70 & 7.98 & 15.41 \\
        5 & \textbf{14.35} & \textbf{9.84} & \textbf{15.69} \\
        7 & 13.51 & 7.07 & 15.44 \\
        \bottomrule
        \end{tabular}
        }
    \end{center}
    \label{tab:variation_with_k}
\end{table}

\subsubsection{Clustering method.} To check if another clustering algorithm might be better, we
create clusters using different algorithms. From table \ref{tab:clustering_method} we observe that
K-means clustering leads to the best performance. Hierarchical agglomerative clustering also gives
close albeit lower performance.

\begin{table}
    \caption{mAPs (\%) for different clustering methods.}
    \begin{center}
        \resizebox{0.42\textwidth}{!}
        {
        \begin{tabular}{@{}lccc@{}}
        \toprule
        \textbf{Clustering} & \textbf{Full} & \textbf{Rare} & \textbf{Non-Rare} \\
        \textbf{Algorithm} & (600 classes) & (138) & (462) \\
        \midrule
        K means & \textbf{14.35} & \textbf{9.84} & 15.69 \\
        Agglomerative & 14.05 & 7.59 & \textbf{15.98} \\
        Affinity Propagation & 13.49 & 7.53 & 15.28 \\
        \bottomrule
        \end{tabular}
        }
    \end{center}
    \label{tab:clustering_method}
\end{table}

\begin{table}
    \caption{Ablation studies (mAP \%).}
    \begin{center}
        \resizebox{0.4\textwidth}{!}
        {
        \begin{tabular}{@{}lccc@{}}
        \toprule
        \textbf{Setting} & \textbf{Full} & \textbf{Rare} & \textbf{Non-Rare} \\
        & (600 classes) & (138) & (462) \\
        \midrule
        Base & \textbf{14.35} & \textbf{9.84} & \textbf{15.69} \\
        Base $- f_h$ &  12.15 & 4.87 & 14.33 \\
        Base $- f_g$ &  12.43 & 8.02 & 13.75 \\
        Base $- w_h - w_o$ & 12.23  & 5.23 & 14.32 \\
        \bottomrule
        \end{tabular}
        }
    \end{center}
    \label{tab:ablation}
\end{table}

\subsubsection{Importance of features.} Further ablation studies (table \ref{tab:ablation}) show
that removing $f_g$, $f_h$, or semantic word-vectors $w_h, w_o$ from the functional generalization
module leads to a reduction in performance. For example, training the model without the geometric
feature $f_g$ gives an mAP of $12.43\%$ and training the model without $f_h$ in the generalization
module gives an mAP of just $12.15\%$. In particular, the performance for Rare classes is quite low.
This shows that these features are important for detecting Rare HOIs. Note that, removing $w_o$ means
that there is no functional generalization.

\subsection{Dealing with Dataset Bias}
Dataset bias leads to models being biased towards particular classes \cite{torralba2011unbiased}. In
fact, bias in the training dataset is usually amplified by the models \cite{zhao2017men,anne2018women}. Our proposed method can be used as a way to overcome the dataset bias problem. To
illustrate this, we use metrics proposed in \cite{zhao2017men} to quantitatively study model bias.

We consider a set of \texttt{(object,predicate)} pairs $\mathcal{Q} =
\{(o_1,p_1),\dots,(o_2,p_2)\}$. For each pair in $\mathcal{Q}$, we consider two scenarios: (1) the
training set is heavily biased \emph{against} the pair; (2) the training set is heavily biased
\emph{towards} the pair. For generating the training sets for a pair $q_i = \{o_i, p_i\} \in
\mathcal{Q}$, for the first scenario, we remove all training samples containing the pair $q_i$ and
keep all other samples for the object. Similarly, for the second scenario, we remove all training
samples containing $o_i$ except those containing the pair $q_i$. For the pair, $q_i$ the test set
bias is $b_i$ (We adopt the definition of bias from \cite{zhao2017men}. See supplementary material
for more details.). Given two models, the one with bias closer to test set bias is considered
better. We show that our approach of augmenting the dataset brings the model bias closer to the test
set bias. In particular, we consider $\mathcal{Q} = \{\texttt{(horse,ride), (cup,hold)}\}$, such
that $b_1 = 0.275$ and $b_2 = 0.305$.

In the first scenario, baseline models trained on biased datasets have biases $0.124$ and $0.184$
for \texttt{(horse,ride)} and \texttt{(cup,hold)} respectively. Note that these are less than the
test set biases because of the heavy bias against these pairs in their respective training sets.
Next, we train models by augmenting the training sets using our methodology for only one neighbor of
each object. 
Models trained on these new sets have biases $0.130$ and $0.195$. That is, our approach leads to a
reduction in the bias \emph{against} these pairs.

Similarly, for the second scenario, baseline models trained on the biased datasets have biases
$0.498$ and $0.513$ for \texttt{(horse,ride)} and \texttt{(cup,hold)} respectively. Training models
on datasets de-biased by our approach give biases $0.474$ and $0.50$. In this case, our approach
leads to a reduction in the bias \emph{towards} these pairs.

%%%%%%%%%%%%%%%%%%%%%%%%%%%%%%%%%%%%%%%%%%%%%%%CONCLUSION%%%%%%%%%%%%%%%%%%%%%%%%%%%%%%%%%%%%%%%%
\section{Discussion and Conclusion}
\label{sec:conclusion}

We discuss some limitations of the proposed approach. First, we assume that all predicates follow
functional similarities. However, some predicates might only apply to particular objects. For
example, you can \texttt{blow} a cake, but not a donut which is functionally similar to cake. Our
current model does not capture such constraints. Further work can focus on trying to
explicitly incorporate such priors into the model. A related limitation is
the independence assumption on predicates. In fact, some predicates are completely dependent. For
example, \texttt{straddle} usually implies \texttt{sit\_on} for bicycles or horses. However, due to
the in-exhaustive labeling of the datasets, we (and most previous work) ignore this dependence.
Approaches exploiting co-occurrences of predicates can help overcome this problem.

\textbf{Conclusion.} We have presented a way to enhance HOI detection by incorporating the common-sense idea that
human-object interactions look similar for functionally similar objects. Our method is able to
detect previously unseen (zero-shot) human-object relationships. We have provided experimental
validation for our claims and have reported state-of-the-art results for the problem. However, there
are still several issues that need to be solved to advance the understanding of the problem and
improve performance of models.

%%%%%%%%%%%%%%%%%%%%%%%%%%%%%%%%%%%%%%%%ACKNOWLEDGEMENT%%%%%%%%%%%%%%%%%%%%%%%%%%%%%%%%%%%%%%%%%%%%%
\section*{\small{Acknowledgement}}
{\scriptsize
    This project was supported by the Intelligence Advanced Research Projects Activity (IARPA) via
Department of Interior/Interior Business Center (DOI/IBC) contract number D17PC00345 and by  DARPA
via ARO contract number W911NF2020009. The U.S.
Government is authorized to reproduce and distribute reprints for Governmental purposes not
withstanding any copyright annotation thereon.\\
\textbf{Disclaimer}: The views and conclusions contained
herein are those of the authors and should not be interpreted as necessarily representing the
official policies or endorsements, either expressed or implied of IARPA, DARPA, DOI/IBC or the U.S.
Government.}

%{\footnotesize
%{\scriptsize
    \bibliographystyle{aaai}
    \bibliography{./biblio}
%}

\beginsupplement
\section{Supplementary Material}

\subsection{Representative clusters}
We claim that the objects in the same cluster can be considered functionally similar. Representative
clusters are: [`Mug', `Pitcher', `Teapot', `Kettle', `Jug'], and [`Elephant', `Dinosaur', `Cattle',
`Horse', `Giraffe', `Zebra', `Rhinoceros', `Mule', `Camel', `Bull']. Clearly, our clusters contain
functionally similar objects. During training, for augmentation we replace the object in a training
sample by other objects from the same cluster. For example, given a training sample for
\texttt{ride-elephant}, we generate new samples by replacing \textit{elephant} by \textit{horse} or
\textit{camel}.% This is shown in figure 3 in the paper.

\subsection{Performance with COCO Detector}
With the original COCO-trained detector, our
method gives an mAP of 16.96, 11.73, and 18.52\% respectively for Full, Rare and Non-Rare sets (up
from 14.37, 7.83, 16.33\% without functional generalization). This performance improvement in even
more significant due to the use of an order of magnitude fewer parameters than existing approaches.
In addition, the proposed approach could be incorporated with any existing method as shown in
the next section.

\subsection{Bonus Experiment: Visual Model}
\label{sec:bonus}
\begin{figure}[ht]
    \centering
        \includegraphics[scale=0.5]{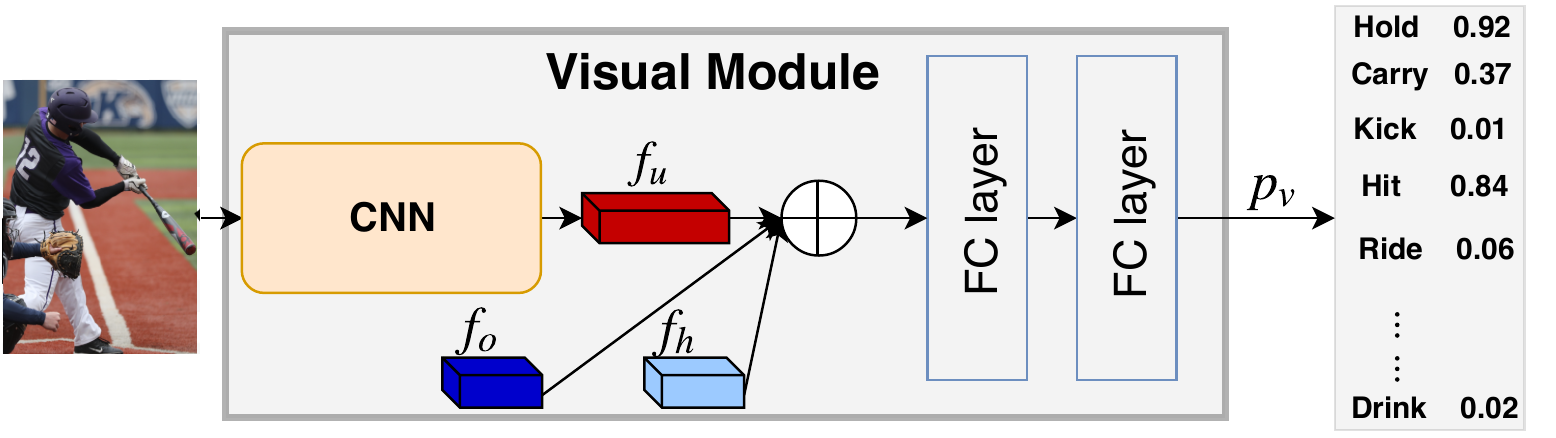}
        \caption{Simple visual module.}
\label{fig:visual}
\end{figure}

Our generalization module can be complementary to existing approaches. To illustrate this, we
consider a simple visual module shown in figure \ref{fig:visual}. It takes the union of $b_h$ and
$b_o$ and crops the union box from the image. It passes the cropped union box through a CNN
(ResNet-50). The feature obtained, $f_u$ is concatenated with $f_h$ and $f_o$ and passed through two FC layers.
This module and the generalization module independently predict the probabilities for predicates and
the final prediction is the average of the two. 
Using the generic object detector, the combined model gives an mAP of $15.82\%$ on the Full HICO-Det
dataset (the visual model separately gives $14.11\%$).
This experiment shows that functional generalization proposed in this paper is complementary to
existing works which rely on purely visual data. Using our generalization module in conjugation with
other existing methods can lead to performance improvements.

\begin{figure*}[h]
    \centering
        \includegraphics[scale=0.24]{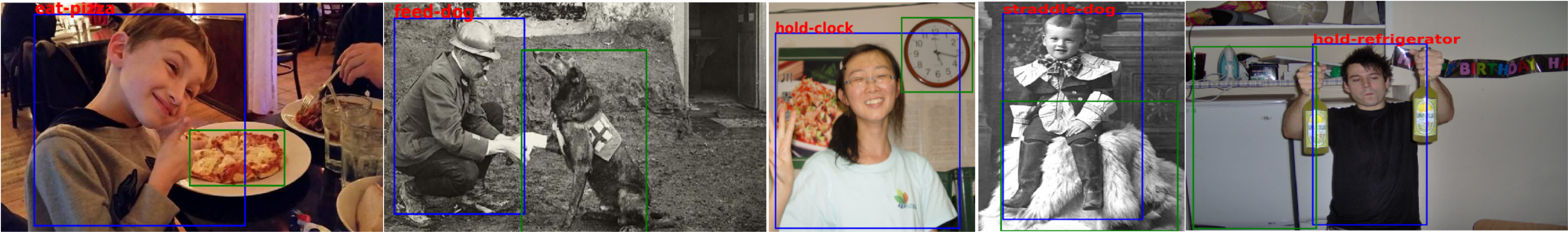}
        \caption{Some incorrect HOI detections in the unseen object ZSD setting. Our model has not seen any
        image with the objects shown above during training.}
    \label{fig:failure_examples}
\end{figure*}

\subsection{Assumptions about number of parameters}
%To study the efficiency of each method, we compute the number of parameters introduced
%by the corresponding method. 
Some works \cite{gupta2018no} have all the details necessary for the
computation in their manuscript, while some
\cite{gao2018ican,li2018transferable,peyre2018detecting} fail to mention the specifics. Hence, we
made the following assumptions while estimating the number of parameters. Note that only those
methods, where sufficient details weren't mentioned in the paper, are discussed.
Since all of the methods use an object detector in the first step, we compute the number of
parameters introduced by the detector. Table \ref{tab:det_params} shows the number of parameters
estimated for each method.

\begin{table}[h]
    \caption{Estimated parameters (in millions) for the detectors used in a few of the
    state-of-the-art methods. (``R-" stands for ``ResNet")}
    \begin{center}
        \resizebox{0.47\textwidth}{!}
        {
        \begin{tabular}{c|c|c}
        \hline
        \textbf{Method} & \textbf{Detector} & \textbf{Params}\\
        \hline
        \cite{gao2018ican} & FPN R-50 &  40.9\\
        \cite{gupta2018no} & Faster-RCNN R-152 &  63.7\\
        \cite{li2018transferable} & Faster-RCNN R-50 &  29\\
        \cite{peyre2018detecting} & FPN R-50 & 40.9 \\
        Ours & Faster-RCNN R-101 & 48 \\

        \hline
        \end{tabular}
        }
    \end{center}
    \label{tab:det_params}
\end{table}
% Mention Detector here
% Backbones
% ResNet50 - 25.5M
% ResNet101 - 44.5M
% ResNet152 - 60.2M                   
% Detectors
% Res50 - 29M
% Res101 - 48M
% Res152 - 63.7M
% Res50-FPN - 40.9M

\subsubsection{ICAN.} % ican - 89.0
% Used a Res50-FPN Backbone
Authors in \cite{gao2018ican} use two fully connected layers in each of the human, object, and
pairwise streams, but the details of the hidden layers were not mentioned in their work. The feature
dimensions of the human and object stream are $3072$, while for the pairwise stream it is $5408$. To
make a conservative estimate, we assume the dimensions of the hidden layers to be $1024$ and $512$ for
the human and object stream. For the pairwise stream we assume dimensions of $2048$ and $512$ for the
hidden layers. We end up with an estimated total of $48.1$M parameters for their architecture. This
gives the total parameters for their method to be \textbf{89}M ($48.1 + $ $40.9$ (Detector; see
table \ref{tab:det_params})).

\subsubsection{Interactiveness Prior.} % Interactiveness prior - 64.0
% Res50 backbone
Li \emph{et al.} \cite{li2018transferable} used a FasterRCNN \cite{ren2015faster} based detector with a
ResNet-50 backbone architecture. In their proposed approach, they have $10$ MLPs (multi-layer
perceptrons) with two layers each and $3$ fully connected (FC) layers. Out of the $10$ MLPs, we
estimated $6$ of them to have an input dimension of $2048$, $3$ of them to have $1024$ and one of
them $3072$. The dimension of hidden layers was given to be $1024$ for all the $10$ MLPs. The $3$ FC layers have input dimensions of $1024$
and an output dimension 117. This gives the number of parameters utilized as $35$M. Their
total number of parameters $=$ \textbf{64}M ($35$ $+$ $ 29$ (detector)).

\subsubsection{Peyre \emph{et al.}} % peyre et al - 65.1
% Res50-FPN backbone

Peyre \emph{et al.} used a FPN \cite{lin2017feature} detector with a ResNet-50 backbone. They have a total
of $9$ MLPs with two hidden layers each, and $3$ FC layers. The input
dimension of the FC layers is $2048$ and the output dimension is $300$. $6$ of the $9$ MLPs have an
input dimension of $300$ and an output dimension of $1024$.  Another $2$ of the $9$ MLPs have input
dimension of $1000$ and $900$ respectively. Their output dimension is $1024$. We assume the
dimensions of the hidden layers in all these MLPs to be $1024$ and $1024$. The last of the $9$ MLPs
has an input dimension of $8$ and an output dimension of $400$. We assume a hidden layer of
dimension $256$ for this MLP. This brings the estimated parameter used to $21.8$M and their  total
parameter count $=$ \textbf{62.7}M ($21.8$ + $ 40.9$ (detector)).

\subsection{Failure cases}
Figure \ref{fig:failure_examples} shows some incorrect detections made by our model in the unseen
object zero-shot scenario. Most of these incorrect detections are very close to being correct. For
example, in the first image, it's very difficult, even for humans to figure out that the person is
not eating the pizza on the plate. In the third and last images, the persons \emph{are} holding
something, just not the object under consideration. Our current model, cannot ignore other objects
present in the scene which lie very close to the person or the object of interest. This is an
area for further research.

\subsection{Bias details}
Adopting the bias metric from \cite{zhao2017men}, we define the bias for a verb-object pair, $(v_*,
o)$ in a set as:
\begin{equation}
    b_s(v_*, o) = \frac{c_s(v_*,o)}{\sum_{v} c_s(v,o)}
\end{equation}
where, $c_s(v,o)$ is the number of instances of the pair $(v,o)$ in the set, $s$. This measure can be used
to quantify the bias for a verb-object pair in a dataset or for a model's prediction. For a dataset,
$\mathcal{D}$, $c_{\mathcal{D}}(v,o)$ gives the number of instances of $(v,o)$ pairs in it.
Therefore, $b_{\mathcal{D}}$
represents the bias for the pair $(v_*,o)$ in the dataset. A low value ($\approx 0$) of
$b_{\mathcal{D}}$ means
that the set is heavily biased against the pair while a high value ($\approx 1$) means that it is
heavily biased towards the pair.

Similarly, we can define the bias of a model by considering the model's predictions as the dataset
under consideration. For example, suppose that the model under consideration gives the predictions
$\mathcal{P}$ for the dataset $\mathcal{D}$. We can define the model's bias as:
\begin{equation}
    b_{\mathcal{P}}(v_*, o) = \frac{c_{\mathcal{P}}(v_*,o)}{\sum_{v} c_{\mathcal{P}}(v,o)}
\end{equation}
where, $c_{\mathcal{P}}(v,o)$ gives the number of instances of the pair $(v,o)$ in the set of the
model's predictions $\mathcal{P}$.

A perfect model is one whose bias, $b_{\mathcal{P}}(v_*,o)$ is equal to the dataset bias
$b_{\mathcal{D}}(v_*,o)$. However, due to bias amplification \cite{zhao2017men,anne2018women}, most
models will have a higher/lower bias than the test dataset depending on the training set bias. That is,
if the training set is heavily biased towards (resp. against) a pair, then the model's predictions
will be more heavily biased towards (resp. against) that pair for the test set. The aim of a bias
reduction method should be to bring the model's bias closer to the test set bias. Our experiments in
the paper showed that our proposed algorithm is able to reduce the gap between the test set
bias and the model prediction bias.

\end{document}